\documentclass[runningheads]{llncs}
\usepackage{graphicx}

\usepackage{tikz}
\usepackage{comment}
\usepackage{amsmath,amssymb} 
\usepackage{color}
\usepackage{float}

\usepackage{marvosym}
\usepackage[symbol]{footmisc}

\usepackage[accsupp]{axessibility}  
\usepackage{multirow}

\usepackage{colortbl}
\definecolor{myy}{RGB}{126,95,0}
\definecolor{mygray}{gray}{.9}
\definecolor{bblue}{RGB}{30,80,120}
\definecolor{mygray1}{gray}{.7}

\definecolor{ggray}{RGB}{127,127,127}
\definecolor{mygreen}{RGB}{93,174,86}
\definecolor{myred}{RGB}{192,0,0}

\usepackage{color, colortbl}
\definecolor{LightCyan}{rgb}{0.88,1,1}
\definecolor{LightRed}{rgb}{1,0.88,0.95}

\newcommand{\tabincell}[2]{\begin{tabular}{@{}#1@{}}#2\end{tabular}}

\begin{document}

\pagestyle{headings}
\mainmatter

\title{MPPNet: Multi-Frame Feature Intertwining with Proxy Points for 3D Temporal Object Detection} %

\titlerunning{MPPNet}
\author{Xuesong Chen\textsuperscript{1,4}\thanks{Equal contributions} \and
Shaoshuai Shi\textsuperscript{2}\protect\footnotemark[1]\thanks{Corresponding authors} \and
Benjin Zhu\textsuperscript{1}  \and 
Ka Chun, Cheung \textsuperscript{3}  \and \\
Hang Xu\textsuperscript{4} \and
Hongsheng Li\inst{1}\protect\footnotemark[2] \\
\protect \textsuperscript{1}MMLab, CUHK \quad
\protect \textsuperscript{2}MPI-INF \quad
\protect \textsuperscript{3}HKBU \quad
\protect \textsuperscript{4}Huawei Noah's Ark Lab\qquad \\
{\tt\small \{chenxuesong@link, hsli@ee\}.cuhk.edu.hk, shaoshuaics@gmail.com}
}

\authorrunning{Chen et al.}
\institute{}
\maketitle

\begin{abstract}
Accurate and reliable 3D detection is vital for many applications including autonomous driving vehicles and service robots. 
In this paper, we present a flexible and high-performance 3D detection framework, named MPPNet, for 3D temporal object detection with point cloud sequences. 
We propose a novel three-hierarchy framework with proxy points for multi-frame feature encoding and interactions to achieve better detection.
The three hierarchies conduct per-frame feature encoding, short-clip feature fusion, and whole-sequence feature aggregation, respectively. To enable processing long-sequence point clouds with reasonable computational resources, intra-group feature mixing and inter-group feature attention are proposed to form the second and third feature encoding hierarchies, which are recurrently applied for aggregating multi-frame trajectory features.
The proxy points not only act as consistent object representations for each frame, 
but also serve as the courier to facilitate feature interaction between frames.
The experiments on large Waymo Open dataset show that our approach outperforms state-of-the-art methods with large margins when applied to both short (\emph{e.g.}, 4-frame) and long (\emph{e.g.}, 16-frame) point cloud sequences. Code is available at \textcolor{red}{https://github.com/open-mmlab/OpenPCDet}.

\end{abstract}

\section{Introduction}
3D object detection from point clouds is an important task for 3D scene perception, which aims to output the 3D bounding boxes and semantic labels of the objects in the given scenes. 
The accurate and reliable 3D detection algorithms are essential for many applications such as the service robots and autonomous driving vehicles, which need accurate detection results for behaviour planning.

However, as one of the most widely adopted depth sensors, LiDAR sensors can only produce point clouds to capture one partial view of the scene at a time. This characteristic leads to incomplete point distributions of objects in driving scenes, posing difficulties for detection methods to accurately estimate the states of the objects. 
In real-world scenarios, the sensors continuously generate point clouds over time as the vehicle moves, and the point cloud sequence naturally provides multiple views of  objects, 
which makes it possible to detect challenging objects more accurately by utilizing the sequence.

Recently, several approaches~\cite{hu2021afdetv2,sun2021rsn,yin2021center} have demonstrated that a simple concatenation of multi-frame point clouds can significantly improve the performance over single-frame detection. However, this strategy is only suitable for handling very short sequences (\emph{e.g.}, 2-4 frames), and its performance might even drop when used to process more frames, since the fused point clouds of long sequences might show ``tails'' (see Fig.~\ref{fig:decoupled_encoding} (a))
of different visual patterns from different moving objects, posing additional challenges to the detectors.
As shown in Table~\ref{table:pilot}, our pilot study demonstrates that CenterPoint~\cite{yin2021center} detector achieves the best performance with 4-frame sequences while it fails on handling longer sequences.
Hence, how to consistently improve the performance of 3D detection with multi-frame point clouds is the main challenge we aim to address in this paper.

\begin{table}[t]
\centering
\caption{Pilot experiments of CenterPoint~\cite{yin2021center} on Waymo validation set in terms of mAPH (LEVEL 2) of the vehicle class by taking concatenated point clouds as input.}
\label{table:pilot}
\resizebox{0.7\textwidth}{!}{
\begin{tabular}{l|ccccc}
\hline \rowcolor{mygray}
Frames & 1-frame & 4-frame  & 8-frame & 12-frame  & 16-frame      \\  \hline
mAPH@L2 &64.50  & \textbf{65.77}  &  65.69     & 65.33   & 64.69     \\ \hline
\end{tabular}
}
\end{table}

We propose a novel two-stage 3D detection framework, named MPPNet, to effectively integrate features from \textbf{M}ulti-frame point clouds via \textbf{P}roxy \textbf{P}oints for 
achieving more accurate 3D detection results.
The first stage adopts existing single-stage 3D detectors to generate 3D proposal trajectories, and we mainly focus on the second stage that takes a 3D proposal trajectory as input to aggregate multi-frame features in an object-centric manner for estimating more accurate 3D bounding boxes.
The key idea of our approach is a series of inherently aligned proxy points to encode consistent representations of an object over time and a three-hierarchy paradigm for better fusing long-term feature sequences.

Specifically, in the first stage, we adopt existing 3D detectors~\cite{yin2021center,lang2019pointpillars} to generate high-quality 3D proposal boxes, and these boxes of all the frames are associated to generate the proposal trajectory for each object, where the box association is based on the predicted speeds of the object.
The proposal trajectories and their associated object points serve as the input to our second stage.  

However, it is challenging to directly aggregate the object features from the multi-frame object points, as the object points at different frames may have significantly different spatial distributions.
To address this challenge, we propose to adopt a set of inherently aligned proxy points at each frame, which are placed at fixed and consistent relative positions in the proposal box of each frame.
These proxy points make the aggregation of the multi-frame features easier by providing consistent representations at different frames.  
Based on the proxy points, our approach can more effectively aggregate the multi-frame features with a three-hierarchy model for 3D temporal object detection. The first hierarchy learns to encode per-frame features, the second hierarchy conducts feature interaction within each short clip (group), and the final hierarchy propagates the whole-sequence information among all the frames and effectively enhances single-frame representations to achieve more accurate detection.

Specifically, in the first hierarchy, to encode per-frame proposal features, 
we adopt set abstraction~\cite{qi2017pointnet++} to aggregate the object point features to a series of relatively consistent proxy points, which encode the object geometry of each frame. 
However, encoding in each frame's local coordinate system inevitably loses the motion information of the proposal trajectory, which is vital for the temporal detection task.
To properly encode the motion features of per-frame proposals, we propose to separately encode the relative positions between the per-frame proxy points and the latest proposal box, which effectively encodes the motion state of per-frame proposal boxes. 
This decoupling strategy for encoding object geometry and motion features at each frame has been experimentally evidenced to benefit the performance of the multi-frame 3D object detection.

The small number of proxy points are also used for feature interaction between multiple frames. However, it is still unaffordable to build connections among all the proxy points as it might incur large costs to computational resources, especially when dealing with long (e.g., 16-frame) trajectories.
Therefore, we adopt a short-clip grouping strategy to temporally divide each proposal trajectory 
into a small number of groups, where each group contains a short clip of sub-trajectory. 
By utilizing the per-frame proxy points as the intermediary, 
our second hierarchy conducts intra-group feature mixing with a novel 3D MLP Mixer module and the third hierarchy further utilizes cross-attention for inter-group feature interactions.
Through this multi-frame feature propagation with our three-hierarchy model, it captures richer trajectory features by aggregating object information from different perspectives of the trajectory, 
which are used to predict a high-quality 3D bounding box by a detection head.

In a nutshell, our contributions are three-fold:
    1) We propose a two-stage 3D detection framework MPPNet,
    which adopts a series of novel proxy points for multi-frame feature encoding and aggregation. Such inherently aligned proxy points not only effectively encode the geometry and motion features of the proposal trajectory in a decoupled manner, but also serve as the courier to facilitate  multi-frame feature interaction.
    2) We present a novel three-hierarchy model to better fuse the multi-frame features of long trajectories, which consists of per-frame encoding, intra-group feature mixing, and inter-group feature attention.  
    3) Our framework can process long point cloud sequences to consistently improve the performance of 3D object detection. Experiments demonstrate that our approach outperforms state-of-the-art methods by large margins when applied to both short (e.g., 4-frame) and long (e.g., 16-frame) sequences.

\section{Related Work}
\noindent
\textbf{3D object detection with single-frame point cloud.} The current single-frame 3D detection algorithms can be roughly divided into three categories: point-based, voxel-based, and jointly employing both of them. 
First, the point-based methods~\cite{shi2019pointrcnn,qi2019deep,yang2019std} directly extracts information from the original point clouds. Through some well-designed operations, like set abstraction~\cite{qi2017pointnet++}, the network can perceive the spatial position features of the irregular 3D point clouds to perform 3D object detection. Different from the point-based method, the voxel-based strategy first converts the irregular points into a number of fixed-size spatial voxels. 
Some high-efficiency methods~\cite{yang2018pixor,lang2019pointpillars,wang2020pillar,ge2020afdet} employ the bird-eye view (BEV) representation, integrating the height information, to encode the voxel features, and then utilize 2D CNN for efficient 3D detection. On the other hand, some works~\cite{yan2018second,zhou2018voxelnet,yin2021center} exploit 3D CNN  to directly extract feature of each voxel in 3D space. Besides, some recent works~\cite{shi2020pv_waymo,li2021lidar,shi2021pv} exploit a voxel-based backbone to obtain 3D proposals, and then use point-based strategies~\cite{qi2017pointnet,qi2017pointnet++} for box refinement, achieving state-of-the-art detection results.

\noindent
\textbf{2D object detection with image videos.}
Different from the scale-invariant 3D bounding boxes, 2D bounding boxes are highly affected by the relative positions of objects and the camera.  
Therefore, some 2D video detection approaches~\cite{kang2017object,kang2017t,zhu2017flow,wang2018fully,han2020mining,jiao2021new} mainly focus on using the appearance and motion features of the previous frame to align the objects in the current frame.
Previous 2D multi-frame object detection methods mainly use optical flow~\cite{zhu2017deep,zhu2017flow,hetang2017impression}, motion~\cite{feichtenhofer2017detect,bertasius2018object}, LSTM~\cite{xiao2018video} to align and aggregate information from different frames. Recently, many methods~\cite{deng2019relation,wu2019sequence,chen2020memory} employ the popular self-attention layer~\cite{vaswani2017attention} as the relation modules to align the features of the previous frame to the current frame, achieving more robust multi-frame detection results.

\noindent
\textbf{3D object detection with point cloud sequences.}
The earlier multi-frame 3D detectors generally adopt a feature-based strategy to gradually aggregate temporal features with 2D CNN~\cite{luo2018fast} or transformer~\cite{9393615,9438625}. 
Besides that, some state-of-the-art 3D detectors~\cite{hu2021afdetv2,sun2021rsn,yin2021center} have demonstrated that a simple multi-frame point concatenation strategy can already outperform the single-frame setting with remarkable margins by taking a short point cloud sequence. 
But when it comes to longer sequences, this simple strategy might fail due to the challenge of handling various moving patterns of objects. 
3D-MAN~\cite{yang20213d} employs attention mechanism  to align different views from 3D object in point cloud and exploits a memory bank to store and aggregate the temporal information to process long point cloud sequence.
SimTrack~\cite{luo2021exploring} presents an end-to-end trainable model for joint detection and tracking from raw point clouds by feature alignment. 
Recently, Offboard3D~\cite{qi2021offboard} proposes an offboard 3D detector that greatly improves the detection performance by taking the whole point cloud sequence as input, but it depends on both past and future frames for generating accurate 3D bounding boxes.
In this paper, we propose a two-stage 3D detector, MPPNet, where we carefully design the multi-frame feature encoding and interaction modules.

\section{Methodology}

Most state-of-the-art 3D detection approaches adopt a simple concatenation strategy~\cite{yin2021center,shi2020pv_waymo}  to take multi-frame point clouds as input for improving 3D temporal detection, which are generally effective at handling short point cloud sequences but fail to deal with long sequences due to the challenge of point-cloud trajectories with different moving patterns. 
3D-MAN~\cite{yang20213d} utilizes the attention networks to aggregate multi-frame features, which, however, may attend to mismatched proposals due to the dense connections among all proposals.

We propose MPPNet, a two-stage 3D detection framework, which takes the point cloud sequences as input to greatly improve the detection results. In the first stage, we adopt existing single-stage 3D  detectors to generate 3D proposal trajectories. Our main innovation focuses on the second stage that takes a 3D proposal trajectory as input and effectively aggregates the multi-frame object features for predicting more accurate 3D bounding boxes. 
In Sec.~\ref{sec:3.1}, we briefly introduce the 1st-stage single-frame proposal generation network and the strategy to generate 3D proposal trajectories. 
In Sec.~\ref{sec:3.2}, we present the 2nd-stage trajectory proposal feature encoding that adopts the proxy point strategy and a three-hierarchy model for multi-frame feature encoding and interaction.
Finally, in Sec.~\ref{sec:3.3}, we introduce the 3D detection head to generate 3D boxes from the summarized temporal features and also discuss the optimization of our approach.

\begin{figure*}[!t]
\centering
\includegraphics[height=6.0cm,width=1.00\textwidth ]{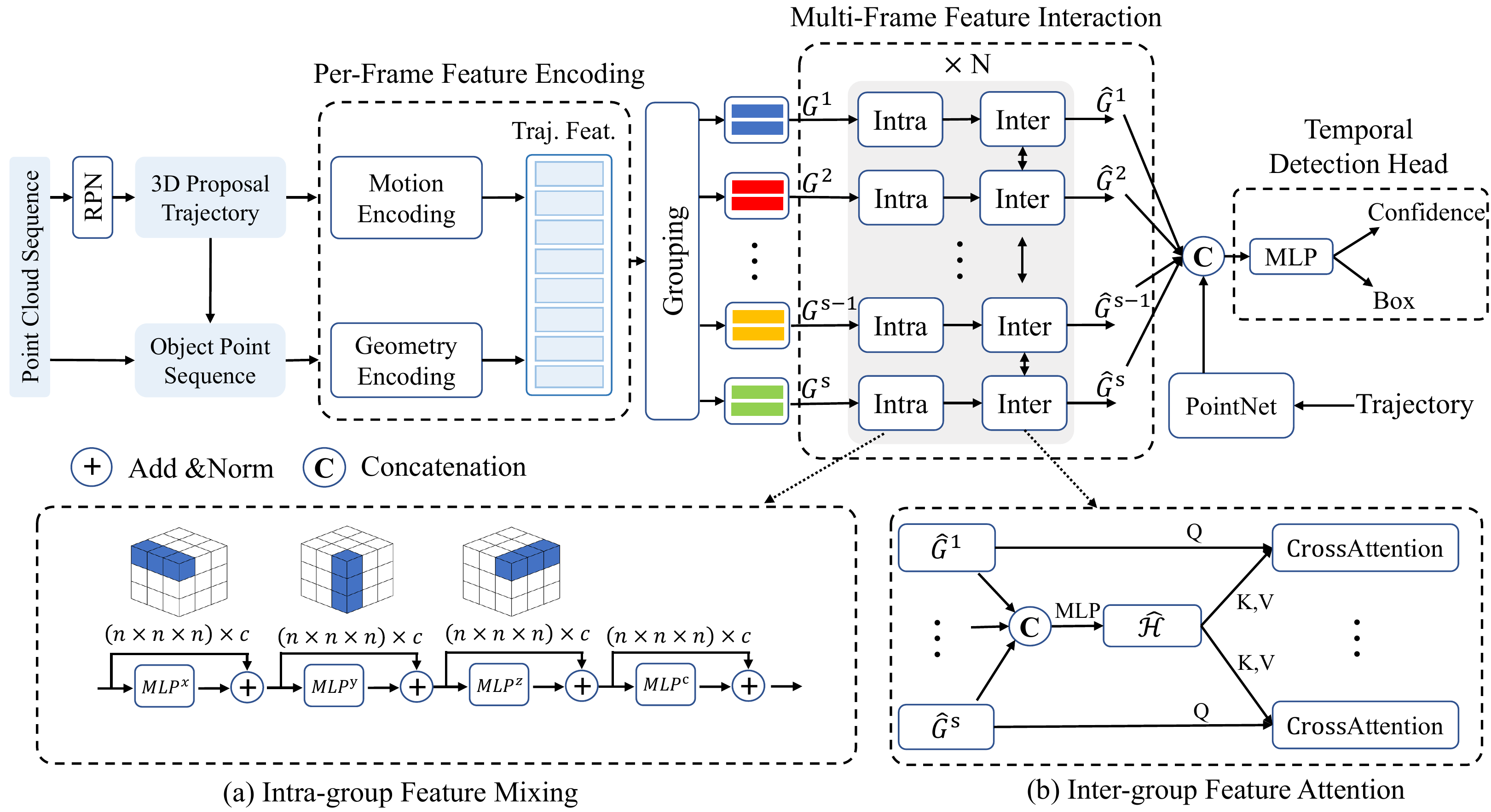}
\caption{The overall architecture of our proposed MPPNet. Our approach takes a point cloud sequence as input for temporal 3D object detection, where a three-hierarchy model is proposed for the multi-frame feature encoding and interaction.}
\label{pipeline}
\end{figure*}

\subsection{Single-frame Proposal Network and 3D Proposal Trajectories}\label{sec:3.1}
In this section, we briefly introduce the 1st-stage of our approach that aims to generate the 3D proposal trajectories. 
Given a point cloud sequence, we create consecutive 4-frame clips and extend the single-stage 3D detectors~\cite{lang2019pointpillars,yin2021center}
as~\cite{hu2021afdetv2,sun2021rsn} to generate per-frame 3D detection boxes. The choice of 4-frame clips is determined according to the results of Table~\ref{table:pilot}.

To associate the 3D detection boxes as 3D proposal trajectories, we add an extra speed prediction head for estimating the speeds of the detected objects, where the speeds are utilized to associate the 3D proposal boxes with a predefined Intersection-over-Union (IoU) threshold as in \cite{yin2021center}.

After associating per-frame 3D proposal boxes to create a 3D proposal trajectory, let $\{\mathcal{B}^1, \dots, \mathcal{B}^T\}$ denote the $T$-frame 3D proposal trajectory and $\mathcal{K}^t=\{l_1^t, \dots,l_m^t\}_{t=1,\cdots, T}$ 
denote object points region pooled~\cite{shi2019pointrcnn} from each frame $t$. They are input into the second stage of our framework for multi-frame feature encoding and interaction to generate more accurate a 3D bounding box.

\subsection{Three-Hierarchy Feature Aggregation with Proxy Points}\label{sec:3.2}
The proposal trajectory and its contained object points capture an object candidate from multiple views,  which provide richer information to estimate its 3D bounding box more accurately. However, it is challenging to aggregate useful information from a long sequence of object points, as the number of object points is large and they also have very different spatial distributions over time.  

To address the challenge, we propose to adopt a set of inherently aligned proxy points at each frame, which not only provide consistent per-frame representations for the object points, but also facilitate multi-frame feature interaction by serving as the courier for cross-time propagation.
With the proxy points, we further present a three-hierarchy model to effectively aggregate multi-frame features from the object point sequence for improve the performance of 3D detection, where the first hierarchy encodes the per-frame object geometry and motion features in a decoupled manner, the second hierarchy performs the feature mixing within each short clip (group) and the third hierarchy propagates whole-sequence information among all clips. By alternatively stacking the intra-group and inter-group feature interactions, the multi-frame features can be well summarized to achieve more accurate detection.

\noindent {\bf Proxy Points.}~
The proxy points are placed at fixed and consistent relative positions in each 3D proposal box of the proposal trajectory. Specifically, at each time $t$, $N=n\times n\times n$ proxy points are uniformly sampled within the 3D proposal box and are denoted as $P^t=\{p_1^t, p_2^t, \dots, p_{N}^t\}$.
These proxy points maintain the temporal order as the order of the trajectory boxes. As they are uniformly sampled in the same way across time, they naturally align the same spatial parts of the object proposals over time. 

\begin{figure*}[!t]
\centering
\includegraphics[height=4.2cm,width=0.95\textwidth ]{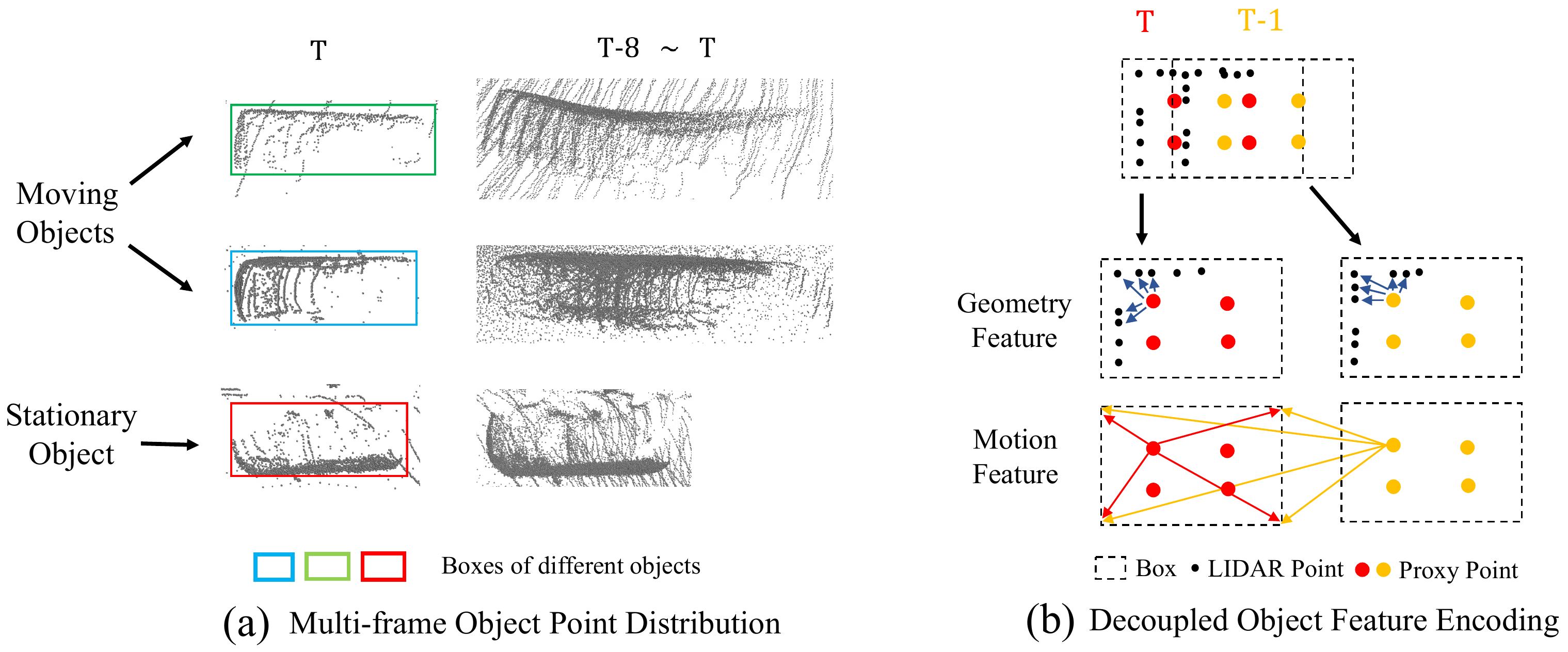}
\caption{Illustration of point distribution of different objects (a) and the decoupled feature encoding for object geometry and motion state (b).}
\label{fig:decoupled_encoding}
\end{figure*}

\noindent
{\bf Decoupled per-frame feature encoding.}
Our first hierarchy performs per-frame feature encoding. The object geometry and motion features are separately encoded based on the proxy points as illustrated in Fig.~\ref{fig:decoupled_encoding} (b). 

For encoding geometry features with proxy points,
the relative differences between each object point $l_i^t\in \mathcal{K}^t$ and the 8 corner + 1 center  points of the proposal box $\mathcal{B}^t = \{b_j^t\}_{j=1}^9$ are first calculated as
\begin{align}\label{eq:point_encoding}
    \bigtriangleup l_{i}^{t} = \text{Concat}\left(\{l_{i}^{t} - b_{j}^t\}_{j=1}^{9}\right).
\end{align}
Hence, the object points' geometry representations are first enhanced as $F^t=\{(l_i^t, \Delta_i^t)\}_{i=1}^m$.
To encode the object geometry features via the proxy points, the set abstraction~\cite{qi2017pointnet++} is adopted to aggregate the neighboring object points to every proxy point $p_k^t$ as
\begin{align}
    {g}_{k}^{t} =\text{SetAbstraction}(p_{k}^{t}, {F^t}), \text{ for } k=1 ,\dots, N.
\end{align}
However, the above geometry feature encoding conducted in the local coordinate of $\{\mathcal{B}^t\}_{t=1}^T$, which inevitably loses the object points' motion information. 

The per-frame motion information is also important for the object points and also needs to be properly encoded. We propose to separately calculate the relative positions between the per-frame proxy points $P^t$ and the latest proposal box's 8 corner + 1 center points $\{b_j^T\}_{j=1}^9$ as Eq.~\eqref{eq:point_encoding}:
\begin{align}
    \Delta p_{k}^{t} =\text{Concat}\left(\{p_{k}^{t} - b_{j}^T\}_{j=1}^{9} \right).
\end{align}
Here, note that the per-frame proxy point $p_k^t$ and the box point $b_j^T$ are all transformed to the same world coordinate system to correctly capture the global motion information.
This strategy effectively encodes the motion state of each proposal box $\mathcal{B}^t$ with respect to the latest box ${\cal B}^T$, as the proxy points at each time $t$ have fixed relative positions in the proposal boxes. 

With an extra one-dimensional time offset embedding $e^t$, 
the final motion encoding of each proxy point is formulated as 
\begin{align}
    {f}_{k}^{t} = \text{MLP}({\rm Concat}(\Delta p_{k}^t, e^t)), \text{ for } k = 1,\dots, N.
\end{align}
By adding the per-frame object geometry features and the motion features, the features of the each proxy point is calculated as
\begin{align}
r_{k}^{t} = {g}_{k}^{t}+ {f}_{k}^{t},  \text{ for } k = 1,\dots, N,
\end{align} 
which forms the final per-frame object features $R^{t} = \{ r_{1}^{t},\dots,r_{N}^{t} \}$ at time $t$. 
\noindent
{\bf Grouping for multi-frame feature interaction.}
The above per-frame encoding aggregates per-frame information to the proxy points, which serve as the representations of each frame to facilitate the multi-frame feature interaction.
However, naively establishing dense connections among all frames' proxy points is still unaffordable and unscaleable in consideration of the tremendous computational costs and GPU memory. 
Therefore, we adopt a grouping strategy to temporally divide the long proposal trajectory into a small number of non-overlapping groups, where each group contains a short sub-trajectory. 

Specifically, the trajectory's per-frame features $\{R^{1}, \dots, R^{T}\}$ are evenly divided into $S$ groups  and each group has $T'$ frames (we assume that $T=T' \times S$). The proposed grouping strategy enables the following multi-frame feature interaction to be more efficiently conducted with the proposed intra-group feature mixing and inter-group cross-attention. 

\noindent
{\bf Intra-group feature mixing.}
The second hierarchy aims to encode group-wise temporal features of the proposal trajectory by conducting the feature interactions within each group. 
We first collect the features of the inherently aligned per-frame proxy points within each group $i$ as
$G^i=\{r_1^{t}, \dots, r_N^{t}\}_{t\in {\cal G}_i}$, where ${\cal G}_i$ is the temporal index set indicating which frames are included in the $i$-th group (e.g., ${\cal G}_1 = \{1,2,3,4\}$ or $\{1,5,9,13\}$ when $S=4$).
The group feature $G^i$ can be represented as a three-dimensional feature matrix that has $G^i\in \mathbb{R}^{T'\times N\times D}$, where $D$ is the feature dimension of each proxy point.
This feature matrix $G^i$ can be considered as an ${N\times (T'\times D)}$ matrix and input to an MLP to obtain group-level proxy points as 
$\hat{G}^{i} = \text{MLP}\left(G^i\right),$
where $\hat{G}^i \in \mathbb{R}^{N\times D}$, and the $\text{MLP}(\cdot)$ fuses each group's representation from $(T'\times D)$ to $D$ dimensions.

To further conduct the feature interaction among all proxy points within each group, inspired by \cite{tolstikhin2021mlp}, we propose the 3D MLP Mixer module to mix each group $i$'s features in the $x,y,z$ and channel dimensions separately and sequentially. 
As illustrated in Fig.~\ref{pipeline} (a), we conduct the feature interaction for each $\hat{G}^{i}$ as
\begin{align}
\hat{G}^{i}= \mathrm{MLP}^{4d}(\hat{G}^{i}), \text{ for } s = 1, \dots, S,
\end{align}
where $\mathrm{MLP}^{4d}(\cdot)$ indicates an MLP with four axis-aligned projection layers for feature mixing along the spatial $x,y,z$ axes and along the channel axis ($c$) on the proxy points, respectively.

\noindent
{\bf Inter-group feature attention.}
After the intra-group feature mixing, our third hierarchy aims to propagate information of the group-level proxy points across different groups to capture richer whole-sequence information. 
We exploit the cross-attention per-group feature representations by querying from the features of the all-group summarization of the whole sequence.

Specifically, the all-group features $\{\hat{G}^{1}, \dots, \hat{G}^{S}\}$ can be represented as a three-dimensional feature matrix denoted as $\mathcal{H} \in \mathbb{R}^{S\times N\times D}$.
It can be considered as an $N\times (S\times D)$ matrix and input to an MLP to obtain the all-group summarization as $\hat{\mathcal{H}} = \text{MLP}\left(\mathcal{H}\right)$, where $\hat{\mathcal{H}} \in \mathbb{R}^{N\times D}$ and the $\text{MLP}(\cdot)$ fuses the all-group summarization from $(S\times D)$ to $D$ dimensions.

Using the all-group summarization as a intermediary, we conduct the inter-group cross-attention for each group $\hat{G}^i$ to aggregate features from $\hat{\mathcal{H}}$ as
\begin{align}\label{eq:attn}
\setlength\belowdisplayskip{9pt}
    \hat{G}^i= \text{MultiHeadAttn} (Q(\hat{G}^i + {\rm PE}), K(\hat{\mathcal{H}} + {\rm PE}), V(\hat{\mathcal{H}})) , \text{ for } i =1,\dots, S,
\end{align}
where $Q(\cdot), K(\cdot), V(\cdot)$ are linear projection layers to generate query, key, value features for the multi-head cross-attention~\cite{vaswani2017attention}, and `$\rm PE$' represents a learnable index-based 3D positional embedding for each proxy point, obtained by projecting proxy points' indices $(i, j, k)$ via a MLP.

The above intra-group feature mixing and inter-group cross-attention are recurrently stacked for multiple times so that the trajectory's representations gradually become aware of both global and local contexts 
for predicting a more accurate 3D bounding box from the trajectory proposal. 

\subsection{Temporal 3D Detection Head and Optimization}\label{sec:3.3}
Through the proposed three-hierarchy feature aggregation, our model obtains richer and more reliable feature representation from input point cloud sequences. 
A 3D detection head to generate final bounding boxes is introduced on top of the above trajectory temporal features, which is supervised by the training losses.

\noindent
{\bf 3D temporal detection head with transformer.}
Given the aforementioned group feature $\hat{G}^i$, we propose to adopt a simple transformer layer to obtain a single feature vector from each group. 
Specifically, we create a learnable feature embedding $E\in \mathbb{R}^{1\times D}$ as the query, to aggregate features from each group feature $\hat{G}^i$ with a multi-head attention as
\begin{align}
    E^i= \text{MultiHeadAttn} ( Q(E), ~K(\hat{G}^i + {\rm PE}), ~V(\hat{G}^i) ) , \text{for } i=1,\dots, S,
\end{align}
where `PE' is similarly defined as that in Eq. \eqref{eq:attn}.
In addition to group features extracted from the raw point clouds, following~\cite{qi2021offboard}, we also exploit a PointNet~\cite{qi2017pointnet} that takes 
$\{\mathcal{B}_1, \dots, \mathcal{B}_T\}$ 
as input to extract the boxes' embedding, where  each box is treated as a point with 7-dim geometry and 1-dim time encoding.

Our detection head therefore integrates group-wise features $\{E^1, \dots, E^S\}$ from both the object points and the boxes' embedding via feature concatenation, for the final confidence prediction and box regression.

\noindent
{\bf Training losses.}
The overall training loss is the summation of the confidence prediction loss $\mathcal{L}_{\mathrm{conf}}$, the box regression loss $\mathcal{L}_{\mathrm{reg}}$ as:
\begin{equation}
\mathcal{L}=\mathcal{L}_{\mathrm{conf}}+ \alpha \mathcal{L}_{\mathrm{reg}}, 
\label{eq:loss}
\end{equation}
where $\alpha$ is the hyper-parameter for balancing different losses.
We adopt the same binary cross entropy loss and box regression loss, employed in CT3D~\cite{sheng2021improving}, as our $\mathcal{L}_{\mathrm{conf}}$ and $\mathcal{L}_{\mathrm{reg}}$.
At training stage, we use the intermediate supervision by adding loss to the intra-group output of each iteration and sum all the intermediate losses to train the model. At test time, we only use the bounding boxes and confidences results predicted from the last intra-group output feature.

\section{Experiments}
\subsection{Dataset and Implementation Details}
\noindent
\textbf{Waymo Open Dataset.} 
The Waymo dataset~\cite{sun2020scalability} is a large-scale 3D detection dataset in autonomous  driving scenarios, which contains 1150 sequences divided into 798 training, 202 validation, and 150 testing sequences. All frames of each sequence are well-calibrated so the 3D temporal detection can be performed. 
Our models are trained on the training set and evaluated on both the validation set and testing set. The official evaluation metrics are standard 3D mean Average Precision (mAP) and mAP weighted by heading accuracy (mAPH). Meanwhile, according to the number of inside points for each object, the data is split into two difficulty levels: LELVEL 1 where objects include more than 5 points and LELVEL 2 where objects include at least 1 points.

\noindent
\textbf{Implementation Details.} 
The two stages of our MPPNet are trained separately. 
We follow the official training strategy for our adopted 1st-stage sub-network (RPN)~\cite{yin2021center,lang2019pointpillars}, while the 2nd-stage sub-network is trained with the ADAM optimizer for 6 epochs with an initial learning rate of 0.003 and a batch size of 16. 
We set the IoU matching threshold to 0.5 for generating the propose trajectories in the 1st stage. 
During training, we conduct the proposal-centric box jitter augmentation as PointRCNN~\cite{shi2019pointrcnn} on the per-frame 3D proposal box. 
We randomly sample 128 raw LiDAR points and set $N=64$ for the proxy points of each proposal in a trajectory. 
The feature dimension $D$ of each proxy point is set as 256. 
For the multi-frame feature interaction, we set the number of groups as $S=4$ for different lengths of trajectories (\emph{e.g.}, for 4-frame input, each frame forms a group; for 16-frame input, every 4 frames forms a group).
We alternately conduct the intra-group feature mixing and inter-group feature attention module for 2 times, followed by an extra intra-group module to attach our proposed 3D detection head for predicting the confidence and 3D bounding box. We set $\alpha=2$ for balancing the terms in the overall loss. During inference, we can store objects' feature of past frames in a memory bank to speed up.

\subsection{Main Results of MPPNet on Waymo Open Dataset}

\noindent
\textbf{Waymo Validation Set.} 
Table~\ref{table:waymo_val} shows the result of published state-of-the-art single-frame and multi-frame methods.
Firstly, when compared with state-of-the-art single-frame method PV-RCNN++~\cite{shi2021pv}, our MPPNet, using CenterPoint~\cite{yin2021center} with 4-frame input as 1st-stage, improves the overall 3D mAPH (LEVEL 2) significantly by $5.05\%$, $8.76\%$, $ 4.9\%$ on vehicle, pedestrian and cyclist, respectively, by exploiting 16-frame point cloud sequences as input. Secondly, compared with multi-frame methods, MPPNet improves the overall 3D mAPH (LEVEL 2) of the vehicle class by $7.82\%$ for 3D-MAN~\cite{yang20213d} and $4.12 \%$ for CT3D-MF, where we extend the open-sourced codes of CT3D~\cite{sheng2021improving} to a multi-frame version CT3D-MF by using the point concatenation strategy. 
Considering that CT3D is a high-performance method, where both the features of raw LiDAR point and proposals from region proposal network are employed, we think it is suitable as a baseline to evaluate the temporal impact from multi-frame points and 3D proposals.  

As demonstrated in Table~\ref{table:waymo_val},
MPPNet can effectively integrate temporal information in point cloud sequences to improve detection accuracy. Specifically, MPPNet achieves significantly gains compared to single-frame detectors, which verifies that our approach successfully exploits temporal information for more accurately estimating  the 3D bounding boxes of some difficult cases in single-frame setting.
Moreover, MPPNet still performs much better than state-of-the-art multi-frame 3D detector 3D-MAN~\cite{yang20213d} when utilizing the same input length of point cloud sequences, indicating that our three-hierarchy design can aggregate spatial-temporal information more thoroughly to benefit 3D detection. 

In addition, we also provide the performance of MPPNet with short sequence input. It can be seen that even with 4-frame input, our method still has advantages over both single-frame and multi-frame methods, improving $3.7\%$ over PV-RCNN++~\cite{shi2021pv} and $2.77\%$ over CT3D-MF~\cite{sheng2021improving} in terms of 3D mAPH (LEVEL 2) of the vehicle class. 
Note that 3D-MAN~\cite{yang20213d}, a representative detection algorithm that can handle long sequence inputs, uses a low-performance PointPillar~\cite{lang2019pointpillars} as their 1st-stage single-frame detection model. 
For fair comparison with 3D-MAN, we also adopt a 1st-stage model that has similar performance with theirs then we extend the model with 4-frame input to get the speed for trajectory generation. 
Table~\ref{table:3D-MAN} demonstrates that MPPNet achieves better results than 3D-MAN. 
Our proposed MPPNet greatly improves the L2 mAPH of single-frame RPN by $16.26\%$, which indicates that our MPPNet can extract useful information to benefit the detection even with the noisy input trajectories. Moreover, when employing a higher-performance single-frame detection model, the performance of our MPPNet can be further improved.

\begin{table}[!t]
\centering
\caption{Performance comparison on the validation set of Waymo Open Dataset. $\dag$ indicates the method implemented by us. The Centerpoint with 4-frame input is adopted as MPPNet's 1st-stage.}
\label{table:waymo_val}
\resizebox{0.98\textwidth}{!}{
\begin{tabular}{l|c|cc|cc|cc|cc}
\hline
\rowcolor{mygray}
& & \multicolumn{2}{c|}{ALL (3D APH)} & \multicolumn{2}{c|}{VEH (3D AP/APH)} & \multicolumn{2}{c|}{PED(3D AP/APH)} & \multicolumn{2}{c}{CYC(3D AP/APH)} \\ \cline{3-10} \rowcolor{mygray}
\multirow{-2}{*}{Method}   &    \multirow{-2}{*}{Frames} & L1  & L2 & L1 & L2  & L1 & L2 & L1  & L2  \\ \hline
SECOND~\cite{yan2018second} & 1  & 63.05   & 57.23   & 72.27/71.69   & 63.85/63.33       &  68.70/58.18 & 60.72/51.31         & 60.62/59.28     &58.34/57.05 \\
PointPillar~\cite{lang2019pointpillars}            & 1             & 63.33      & 57.53       & 71.60/71.00  & 63.10/62.50        &  70.60/56.70   & 62.90/50.20     & 64.40/62.30    & 61.90/59.90                 \\
LiDAR R-CNN~\cite{li2021lidar}            & 1                       & 66.20       & 60.10      & 73.50/73.00       & 64.70/64.20      & 71.20/58.70       & 63.10/51.70      & 68.60/66.90      & 66.10/64.40     \\
RSN~\cite{sun2021rsn}                     & 1 & -  & -     & 75.10/74.60         & 66.00/65.50        & 77.80/72.70                 & 68.30/63.70                 & -                 &  -               \\
Pyramid~\cite{mao2021pyramid}          & 1     & - & -                 & 76.30/75.68       & 67.23/66.68  &-  &-  &- &-                 \\
PV-RCNN~\cite{shi2020pv_waymo}             & 1  & 69.63   & 63.33   &77.51/76.89 & 68.98/68.41  & 75.01/65.65  & 66.04/57.61                  & 67.81/66.35   & 65.39/63.98                 \\
Part-A2~\cite{shi2020points}             & 1                      & 70.25  & 63.84   & 77.05/76.51   & 68.47/67.97  & 75.24/66.87     & 66.18/58.62     & 68.60/67.36  & 66.13/64.93         \\
Centerpoint~\cite{yin2021center}             & 1                       & -                 & 65.50       & -  & -/66.20        & -   & -/62.60     & -   & -/67.60                 \\
CT3D~\cite{sheng2021improving}              & 1                      & -                  & -               & -       & 69.04/-     &  -                & -                 &  -                & -                \\
PV-RCNN++~\cite{shi2021pv}                & 1     & 75.21  & 68.61   & 79.10/78.63       & 70.34/69.91 & 80.62/74.62  & 71.86/66.30	 & 73.49/72.38  & 70.70/69.62                 \\ \hline
3D-MAN~\cite{yang20213d}                  & 16                      &-                   &-     & 74.53/74.03  & 67.61/67.14 & -  & -                & -                &  -               \\
$\dag$ Centerpoint~\cite{yin2021center}               & 4    &  74.88     &  69.38                &76.71/76.17       & 69.13/68.63      &  78.88/75.55       &  71.73/68.61                &  73.73/72.96               & 71.63/70.89              \\
$\dag$ CT3D-MF~\cite{sheng2021improving}                & 12    &  -     &  -                &79.30/78.82       & 71.82/70.84      &  -                &  -                &  -                & -                \\
$\dag$ CT3D-MF~\cite{sheng2021improving}                & 16    &  -     &  -                &79.04/78.55       & 71.14/70.68      &  -                &  -                &  -                & -                \\
MPPNet (Ours)     & 4                       & 79.83   & 74.22  & 81.54/81.06 & 74.07/73.61 &84.56/81.94    & 77.20/74.67      & 77.15/76.50     & 75.01/74.38                 \\
MPPNet (Ours)   & 16                      & \textbf{80.40}   & \textbf{74.85}   & \textbf{82.74/82.28} & \textbf{75.41/74.96}  & \textbf{84.69/82.25}   & \textbf{77.43/75.06}      &\textbf{77.28/76.66}     & \textbf{75.13/74.52}  \\ \hline  
\end{tabular}}

\end{table}

\begin{table}[!t]
\centering
\caption{Performance comparison on the testing set of Waymo Open Dataset.}
\label{table:waymo_test}
\resizebox{0.99\textwidth}{!}
{
\begin{tabular}{l|c|cc|cc|cc|cc}
\hline
\multirow{2}{*}{\cellcolor{mygray}} & \multirow{2}{*}{\cellcolor{mygray}} & \multicolumn{2}{c|}{\cellcolor{mygray}ALL (3D APH)} & \multicolumn{2}{c|}{\cellcolor{mygray}VEH (3D AP/APH)} & \multicolumn{2}{c|}{\cellcolor{mygray}PED(3D AP/APH)} & \multicolumn{2}{c}{\cellcolor{mygray}CYC(3D AP/APH)} \\ \cline{3-10} 
    \multirow{-2}{*}{\cellcolor{mygray}Method}          &     \multirow{-2}*{\cellcolor{mygray}Frames}   &  \cellcolor{mygray} L1                & \cellcolor{mygray}L2 & \cellcolor{mygray}L1   & \cellcolor{mygray}L2 & \cellcolor{mygray}L1   & \cellcolor{mygray}L2        & \cellcolor{mygray}L1               & \multirow{-1}*{\cellcolor{mygray}L2}              \\ \hline
StarNet~\cite{ngiam2019starnet}                 & 1                       & -                 & -                & 61.50/61.00         & 54.90/54.50       &  67.80/59.90    & 61.10/54.00   & -  & -                 \\
PointPillar~\cite{lang2019pointpillars}             & 1                       & -                 & -                & 68.60/68.10         & 60.50/60.10     &68.00/55.50    & 61.40/50.10   & -  & -                 \\
CenterPoint~\cite{yin2021center}             & 1                       & -                 & -                & 80.20/79.70         & 72.20/71.80        & 78.30/72.10     &72.20/66.40   & -  & -                 \\

PV-RCNN++~\cite{shi2021pv}    & 1                &75.65    & 70.21    & 81.62/73.86       & 81.20/73.47      & 80.41/74.12  & 74.99/69.00          & 71.93/69.28   & 70.76/68.15    \\ \hline
RSN~\cite{sun2021rsn}                    & 3                       & -                 & -                & 80.70/80.30         & 71.90/71.60    &78.90/75.60     & 70.70/67.80  &  -  &  -               \\
Centerpoint~\cite{yin2021center}              & 2  & 77.18  & 71.93 & 81.05/80.59       & 73.42/72.99      & 80.47/77.28  & 74.56/71.52 & 74.60/73.68                   & 72.17/71.28                \\
PV-RCNN Ens~\cite{shi2020pv_waymo}             & 2         & 76.90   & 71.52    & 81.06/80.57       & 73.69/73.23      & 80.31/76.28      & 73.98/70.16      &75.10/73.84    & 72.38/71.16  \\
Pyramid~\cite{mao2021pyramid}          & 2                       & -                  & -                 & 81.77/81.32       & 74.87/74.43      &-                  & -                 &  -                &-                 \\
3D-MAN~\cite{yang20213d}                  & 16                      & -                  &  -                & 78.71/78.28       & 70.37/69.98      & 69.97/65.98   & 63.98/60.26                 &     -             & -                \\
MPPNet (Ours)     & 16                      & \textbf{80.59}   & \textbf{75.67}  & \textbf{84.27/83.88}   & \textbf{77.29/76.91}    & \textbf{84.12/81.52}                  & \textbf{78.44/75.93}     & \textbf{77.11/76.36}     & \textbf{74.91/74.18}     \\ \hline
\end{tabular}
}
\end{table}

\begin{table}[t]
\begin{minipage}[b]{0.48\linewidth}
\centering
\caption{Performance comparison with 3D-MAN~\cite{yang20213d}, where ES and FT indicate early-stop and fully-trained models.}
\label{table:3D-MAN}
	\resizebox{0.99\textwidth}{!}{
		\renewcommand\arraystretch{1.0}
\begin{tabular}{l|c|l}
\hline
\rowcolor{mygray}
Method          & Frames & mAPH@L2     \\ \hline
PointPillar        & 1      & 54.69         \\
PointPillar-ES      & 1      & 55.13         \\
PointPillar-FT   & 1      & 64.37         \\ \hline
3D-MAN w/ PointPillar           & 16     & 67.14 (+12.45)  \\
Ours w/ PointPillar-ES & 16     & 71.39 (+16.26) \\
Ours w/ PointPillar-FT & 16     & 72.90 (+8.53)   \\ \hline
\end{tabular}}

\end{minipage}
\hspace{10pt}
\begin{minipage}[b]{0.45\linewidth}
	\centering
	\caption{Effects of the input trajectory length. All experiments adopt the CenterPoint~\cite{yin2021center} as RPN by training with 6 epochs for saving the training cost.}
	\label{table:frames}
	\resizebox{0.9\textwidth}{!}{
		\renewcommand\arraystretch{1.0}
\begin{tabular}{c|c|c}
\hline
\rowcolor{mygray}
Method   & \multicolumn{1}{c|}{CT3D-MF}       & \multicolumn{1}{c}{MPPNet}        \\ \hline
4-frame  & 70.19         & 72.63         \\ 
8-frame  & 70.71 (+0.52) & 73.22 (+0.59) \\ 
12-frame & 70.84 (+0.65) & 73.55 (+0.92) \\ 
16-frame & 70.68 (+0.49) & 73.81 (+1.18)  \\ \hline
\end{tabular}}
	
\end{minipage}
\end{table}

\noindent
\textbf{Waymo Testing Set.} 
We also evaluate our method on the testing set of Waymo dataset to further validate the generalization of our approach.
As shown in Table~\ref{table:waymo_test}, 
our approach achieves much better performance than both the state-of-the-art single-frame method~\cite{shi2021pv} and multi-frame methods~\cite{yin2021center,yang20213d} by large margins. 
Note that we do not use test-time augmentations and model ensemble.

\begin{table}[t]
\centering
\caption{Effects of different components in our proposed framework, where the first row is our MPPNet with 4-frame sequences and default settings for comparison.}
\label{table:ablation}
\resizebox{0.98\textwidth}{!}{
\begin{tabular}{c|c|c|cc|c}
\hline
\rowcolor{mygray}
& \multirow{2}{*}{\tabincell{c}{Boxes's \\Embedding}} & & \multicolumn{2}{c|}{Multi-frame Feature Fusion} & \\ \cline{4-5} \cellcolor{mygray}\multirow{-2}{*}{\cellcolor{mygray} Proxy Point} &  &\cellcolor{mygray} \multirow{-2}{*}{\begin{tabular}[c]{@{}c@{}}Per-frame Feature\\ Encoding\end{tabular}}  &  \multicolumn{1}{c}{\cellcolor{mygray} Intra}     & \multicolumn{1}{c|}{\cellcolor{mygray} Inter} &   \cellcolor{mygray}      \multirow{-2}{*}{mAPH@L2}             \\ \hline 
$\checkmark$                 & $\checkmark$           & Geometry+Motion  & \multicolumn{1}{c|}{3D MLP Mixer}     & Cross-Attn & \multicolumn{1}{l}{73.08} \\ \hline 
$\times$               & $\checkmark$           & Geometry+Motion        & \multicolumn{1}{c|}{Self-Attn}        & Cross-Attn  & 71.13 (-1.95)       \\ 
$\checkmark$                 & $\checkmark$           & Geometry+Motion  & \multicolumn{1}{c|}{Self-Attn}        & Cross-Attn  & 72.71 (-0.37)       \\ 
$\checkmark$                 & $\checkmark$           & Geometry         & \multicolumn{1}{c|}{3D MLP Mixer}     & Cross-Attn  & 72.78 (-0.30)       \\ 
$\checkmark$                 & $\checkmark$           & Integrated       & \multicolumn{1}{c|}{3D MLP Mixer}     & Cross-Attn  & 72.89 (-0.19)       \\ 
$\checkmark$                 & $\checkmark$           & Geometry+Motion  & \multicolumn{1}{c|}{3D MLP Mixer}     & $\times$ & 72.36 (-0.72)        \\ 
$\checkmark$                 & $\checkmark$           & Geometry+Motion  & \multicolumn{1}{c|}{3D MLP Mixer}     & Cross-Attn w/o PE  & 72.97 (-0.11)        \\ 
$\checkmark$                 & $\checkmark$           & Geometry+Motion  & \multicolumn{1}{c|}{3D MLP Mixer}     & Cross-Attn w/o Sum. & 72.95 (-0.13)        \\ 
$\checkmark$                 & $\times$         & Geometry+Motion        & \multicolumn{1}{c|}{3D MLP Mixer}     & Cross-Attn & 72.98 (-0.10)        \\ \hline
\end{tabular}}
\end{table}

\subsection{Ablation Studies}
We conduct comprehensive ablation studies to verify the effectiveness of each component of MPPNet. To save the training cost, unless otherwise mentioned, all ablation experiments of MPPNet are trained on the vehicle category for 3 epochs by taking four-frame point cloud as input. We take the mAPH (LEVEL 2) as the default metric for comparison. The results are shown in Table \ref{table:ablation}.

\noindent
\textbf{Effects of the input length of trajectory and boxes' embedding.}
Table~\ref{table:frames} verifies the ability of MPPNet to aggregate multi-frame temporal information from the trajectories with different lengths.
We observe that CT3D-MF~\cite{sheng2021improving} achieves the best performance when processing 12-frame sequences, demonstrating that it has relatively better temporal modeling capabilities than CenterPoint~\cite{yin2021center} but still struggles at handling longer sequences (\emph{e.g.} 16 frames).
In contrast, our MPPNet can constantly improve the performance by taking longer sequences (\emph{i.e.}, from 4 to 16 frames) as input, which demonstrates the effectiveness of our proxy point based three-hierarchy model for multi-frame feature encoding and interaction.
In addition, removing the boxes' embedding leads to a slight performance drop of $-0.1\%$ (see the last row of Table~\ref{table:ablation}).

\noindent
\textbf{Effects of proxy points.}
We investigate the importance of the proposed proxy points by comparing with a baseline that directly use raw LiDAR points. 
Considering that the raw LiDAR points are unordered, we first replace 3D MLP Mixer with a self-attention module for intra-group feature fusion, 
and then we employ 128 LiDAR points' feature as the representation of each proposal for the multi-frame feature interaction. 
The $2^{rd}$ and $3^{th}$ rows of Table~\ref{table:ablation} show that removing the proxy point substantially decreases the performance by $1.95\%$. 
It demonstrates that our proposed proxy points can provide consistent representations for different frames and greatly benefit feature propagation among all the frames, while directly aggregating temporal features from the unordered point clouds is very less effective. Moreover, we explore the effects of different numbers of proxy points in Table~\ref{table:proxypoint_num}. 
Here we observe that a smaller number of proxy points (\emph{i.e.}, $27 = 3^3$) degrades the performance by $-0.54\%$ since the too few proxy points cannot finely represent the geometric details of objects, while adopting more proxy points (\emph{i.e.}, 64 and 125) achieves similar performance but more proxy points lead to more computations. Hence our MPPNet adopts $64=4\times 4\times 4$ proxy points as a trade-off. 

\noindent
\textbf{Effects of intra-group feature mixing.}~
We adopt different designs to investigate the effectiveness of our proposed 3D MLP Mixer for intra-group feature mixing.
As shown in the $3^{rd}$ row of Table~\ref{table:ablation}, we replace the proposed 3D MLP Mixer by a standard self-attention module~\cite{vaswani2017attention}, which results in a $0.37\%$ performance drop compared with the $1^{st}$ row. It demonstrates that the information propagation manner in 3D MLP Mixer is better for intra-group feature fusion.

\noindent
\textbf{Effects of inter-group feature interaction.} 
By comparing $1^{st}$ and $6^{th}$ rows of Table~\ref{table:ablation}, we observe that the performance drops a lot ($-0.72\%$) after removing the inter-group feature attention module.
It demonstrates that the interactions among different groups are essential for achieving accurate detection, since the model can better integrate multi-view information from the whole 3D trajectory.
Moreover, we also provide two variants of the proposed cross-attention module to verify the effectiveness of our design. 
For the first variant, we remove our proposed index-based position embedding in the cross attention, and by comparing $1^{st}$ and $7^{th}$ rows of Table~\ref{table:ablation} we observe that the proposed index-based position embedding brings $0.11\%$ performance gain. 
Meanwhile, in another variant of inter-group cross-attention, we use each group's features as query ($N\times D$) to aggregate features from the concatenation of all other groups' features ($((S-1)\times N)\times D$), rather than using the all-group summarization, and the $1^{st}$ and $8^{th}$ rows of Table~\ref{table:ablation} show that the performance drops a bit (-$0.13\%$).

\noindent
\textbf{Effects of decoupled encoding for geometry and motion Feature.}
As shown in the $4^{th}$ and $5^{th}$ rows of Table~\ref{table:ablation}, without motion encoding, MPPNet suffers from  a performance drop by $-0.3\%$, suggesting that motion information is beneficial for state estimation of objects from point cloud sequences. Meanwhile, compared with our proposed decoupled encoding strategy, the integrated encoding manner by attaching a per-frame time encoding to each LiDAR point will degrade the performance by $-0.19\%$, which proves that the decoupled feature encoding strategy is beneficial for the network's feature learning.
\begin{table}[t]
\begin{minipage}[b]{0.45\linewidth}
	\centering
	\caption{Effects of different numbers of proxy points in our MPPNet. $n\times n\times n$ proxy points are uniformly sampled within each proposal box.
	}
	\label{table:proxypoint_num}
	\resizebox{0.7\textwidth}{!}{
		\renewcommand\arraystretch{1.0}
\begin{tabular}{c|c}
\hline
\rowcolor{mygray}
$\#$ (Proxy Point) & mAPH@L2 \\ \hline 
\text{$3\times3\times3$} & 72.54 \\ 
\text{$4\times4\times4$} & 73.08 \\ 
\text{$5\times5\times5$} & 72.98 \\ \hline 
\end{tabular}
}

\end{minipage}
\hspace{6pt}
\begin{minipage}[b]{0.49\linewidth}
	\centering
	\caption{
	Effects of trajectory augmentation, intermediate supervision and grouping strategy. The upper part investigates the training strategy with 4-frame input, while the lower part explores the grouping strategy with 16-frame input. }
	
	\label{table:train_strategy}
	\resizebox{0.99\textwidth}{!}{
		\renewcommand\arraystretch{1.0}
\begin{tabular}{c|c|c|c}
\hline
\multirow{2}{*}{\begin{tabular}[c]{@{}c@{}}Training \\ Strategy \end{tabular}}& \cellcolor{mygray} MPPNet  & \cellcolor{mygray} w/o  Traj.  Aug & \cellcolor{mygray} w/o Int. Loss \\ \cline{2-4} & 73.08   & 72.62 (-0.46)   & 72.47 (-0.61) \\ \hline
\multirow{2}{*}{\begin{tabular}[c]{@{}c@{}}Grouping\\ Strategy\end{tabular}} & \cellcolor{mygray} Stride 4 &  \cellcolor{mygray} Stride 1         &  \cellcolor{mygray}-             \\ 
\cline{2-4}  & 74.21   & 74.06           & -             \\ \hline
\end{tabular}
}
\end{minipage}
\end{table}

\noindent
\textbf{Effects of trajectory augmentation and intermediate supervision.}
The top part of Table~\ref{table:train_strategy} shows 
that the performance drops by $-0.46\%$ without data augmentation, and drops by $-0.61\%$ without the intermediate supervisions.
We consider that the intermediate loss can help the network to gradually optimize and refine the features by supervising them after each block. 

\noindent
\textbf{Effects of clip grouping strategy.}
We investigate different grouping strategies' impact with 16-frame input. As shown in the bottom part of Table~\ref{table:proxypoint_num}, when setting stride to 4 to form the groups, {\it e.g.}, $\{1,5,9,13\}$ (denoted as `stride 4'), the network achieves better results than using temporal stride 1 to form every group, {\it e.g.}, $\{1,2,3,4\}$ (denoted as `stride 1').

\section{Conclusions}
In this work, we present a two-stage 3D detection framework MPPNet, adopting a series of novel proxy points to integrate multi-frame features for 3D object detection. With the inherently aligned proxy points, 
MPPNet encodes the per-frame object geometry and motion features in an decoupled manner then they are deeply interacted through our proposed intra-group feature mixing and inter-group feature attention, which effectively propagate the features among the per-frame proxy points in the whole trajectory to perform more accurate 3D detection.   
The  experiments on large-scale Waymo Open dataset demonstrate that our approach captures better multi-view features from the point cloud sequences and outperforms state-of-the-art methods  with remarkable margins.

\section{Acknowledgement}
This work is supported in part by NVIDIA, in part by Centre for Perceptual and Interactive Intelligence Limited, in part by the General Research Fund through the Research Grants Council of Hong Kong under Grants (Nos. 14204021, 14207319), in part by CUHK Strategic Fund. Hongsheng Li is also a PI at Centre for Perceptual and Interactive Intelligence Limited.

\bibliographystyle{splncs04}
\bibliography{egbib}
\end{document}